\documentclass[]{fairmeta}

\usepackage[utf8]{inputenc}
\usepackage[T1]{fontenc}
\usepackage{hyperref}
\usepackage{url}
\usepackage{booktabs}
\usepackage{amsfonts}
\usepackage{amsmath}
\usepackage{amssymb}
\usepackage{mathtools}
\usepackage{nicefrac}
\usepackage{microtype}
\usepackage{xcolor}
\usepackage{graphicx}
\usepackage{multirow}
\usepackage{caption}
\usepackage{float}            

\usepackage[textsize=tiny]{todonotes}

\definecolor{linkcolor}{RGB}{84,84,192}
\hypersetup{
    colorlinks=true,
    linkcolor=linkcolor,
    citecolor=linkcolor
}

\newcommand{\model}{DANCE}

\newcommand{\rev}[1]{\textcolor{black}{#1}}

\title{DANCE: Detect and Classify Events in EEG}

\author[1,2]{Jarod Lévy}
\author[1]{Hubert Banville}
\author[1]{Jérémy Rapin}
\author[1]{Jean-Remi King}
\author[2\dagger]{Thomas Moreau}
\author[1\dagger]{Stéphane d'Ascoli}

\affiliation[1]{Meta AI}
\affiliation[2]{Inria, Université Paris-Saclay, Palaiseau, France}

\contribution[\dagger]{These authors jointly supervised this work}

\date{\today}
\correspondence{\email{jarod@meta.com}}

\abstract{
Event identification in continuous neural recordings is a critical task in neuroscience. \rev{Decoding in EEG is dominated by classifying windows aligned to known event onsets. However, while available in controlled experiments, such onsets are absent in continuous real-world monitoring.}
\rev{Here, we introduce \model{}, a deep learning pipeline that frames neural decoding as a set-prediction problem and jointly detects and classifies events directly from raw, unaligned signals.}
Evaluated \rev{separately} on ten datasets curated from the literature with a wide variety of event types ranging from milliseconds to minutes in duration, our model outperforms existing methods on a broad range of cognitive, clinical and BCI tasks. This single architecture establishes a new state of the art in the competitive task of seizure monitoring and matches the accuracy of onset-informed models for BCI tasks. Overall, our method marks a step towards end-to-end \rev{asynchronous neural} decoding models.
}

\begin{document}

\maketitle

\section{Introduction}

% The duplicate abstract that used to live here has been moved to the
% preamble \abstract{...} for the fairmeta class (it is rendered inside
% \maketitle's title box).

% =========================================================================

\begin{figure}[h]
    \centering
    \includegraphics[trim=1cm 1cm 1cm 2.7cm,clip,width=\linewidth]{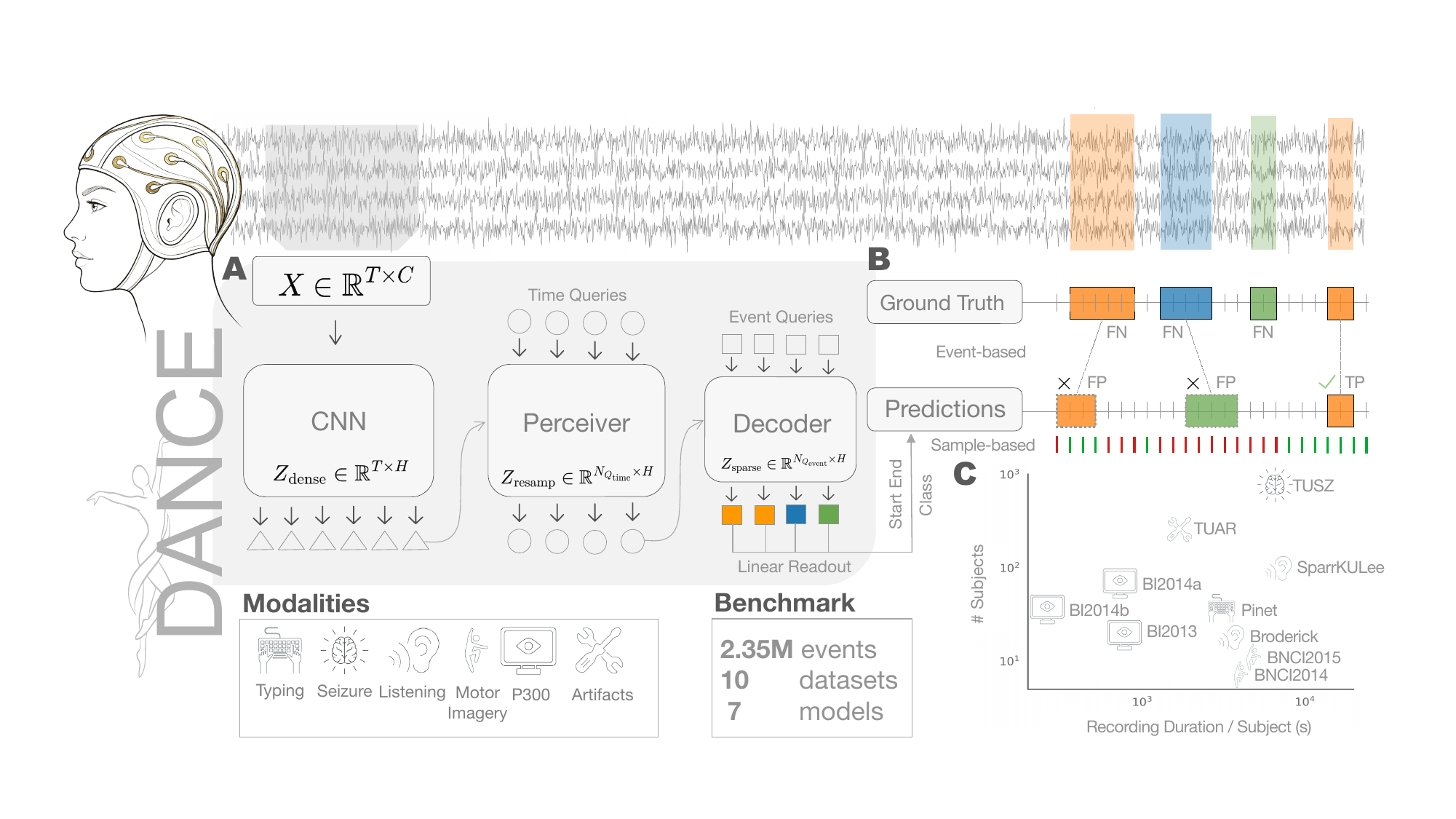}
    \caption{\textbf{Our Approach.}
    \textbf{A.} The \model{} model processes EEG time-series windows of fixed length through a CNN backbone, a Perceiver and a Decoder.
    \textbf{B.} \model{} predicts event instances of different classes (here, represented by different colors), which are evaluated against ground truth using both event-based and sample-based metrics. Dashed contours represent incorrect events according to the event-based metric.
    \textbf{C.} 
    % The model is trained on a large-scale, heterogeneous benchmark comprising 1{,}154 participants and 2.35 million events. 
    \rev{The model is trained separately on 10 heterogeneous datasets comprising a total of 2.35 million events and is compared with 7 different models.}
    The data spans six distinct modalities: Typing, Seizure, Listening (Speech), Motor Imagery, P300 and Artifacts, with varying subject counts and recording volumes.
    }
    \label{fig:approach}
\end{figure}

\section{Introduction}

% Why events matter
Temporal events are the primary unit of analysis in experimental and clinical neuroscience. Experimental datasets typically rely on discrete, time-locked occurrences, such as motor-imagery trials \citep{faller2012, tangermann2012}, visual stimuli \citep{congedo2011, korczowski_a, korczowski_b}, or speech utterances \citep{accou, broderick}. In clinical contexts, this focus on temporally localized phenomena is equally central, for instance in the detection of transient events such as epileptic seizures \citep{shah2018}. \rev{Beyond controlled environments, real-world applications, including real-time neuroprosthetics \citep{willett_handwriting}, continuous wheelchair control \citep{zhang2024brain}, and naturalistic neurofeedback \citep{le2022toward}, require moving from discrete event analysis to asynchronous, continuous decoding.} Bridging these settings therefore hinges on the ability to reliably detect and classify events under increasingly unconstrained conditions. \looseness=-1

% Dominant paradigm: onset-informed classification
Despite this need, decoding in EEG remains constrained by its inherently low signal-to-noise ratio (SNR) \citep{snr_eeg}. To mitigate this, the dominant practice in event classification benchmarks is to evaluate models on pre-segmented windows time-locked to known event onsets, bypassing the detection task to focus purely on classification \citep{moabb, defossez, banville2025scaling, cheng2020subject, saeed2021learning, dcrnn, levy2025, graphs4former}. While useful for controlled comparisons, this onset-informed paradigm fails to reflect realistic settings where precise onset markers are unavailable. As a result, achieving robust asynchronous detection and classification in non-invasive settings remains an open challenge.

% Shift happened in domains with higher SNR
% In domains with higher SNR, such as electromyography (EMG; \citealt{emg2qwerty, splashnet}) or intracranial recordings \citep{willett_handwriting}, recent works have proposed end-to-end formulations using approaches based on Connectionist Temporal Classification (CTC; \citealt{ctc}), e.g.\ for speech decoding \citep{metzger_decoding}. A method that operates effectively under this end-to-end setting is still elusive for EEG.

% Define dense vs sparse + Computer Vision Inspirations
There are two ways to frame event detection problems: as a dense (or \emph{sample-based}) prediction task, where the objective is to predict whether an event is present for each time sample in a window, or as a sparse (or \emph{event-based}) prediction task, where the objective is to predict directly the set of events occurring within the entire window. We illustrate these two types of approaches in \Cref{fig:approach}B. In computer vision, significant advances have been achieved in object detection by switching from the dense view to the sparse view, via the introduction of detection architectures \citep{rcnn, faster_rcnn, yolo} tailored for structured prediction and object-level metrics \citep{maier2024metrics} such as the Intersection over Union (IoU). Prior work reframed detection as a set-prediction problem, most notably with the DETR family of models \citep{carion2020end, dn_detr, lv2023detrs}. Extending this paradigm, the Temporal Action Detection literature introduces hybrid strategies that combine set-based objectives with dense auxiliary supervision \citep{codetr, wang2024rt}, leading to more stable and efficient training. 

\rev{Directly applying computer vision methods to scalp EEG fails due to low SNR, limited per-dataset sample size, and event durations spanning four orders of magnitude (from milliseconds to minutes). The asynchronous neural-decoding literature mostly relies on sliding-window classification. The most straightforward approaches apply a classifier to every window of the signal to get a per-timestep prediction~\citep{craik2019deep}. More efficient methods utilize two-stage cascades first detect event presence, then classify positive windows \citep{wu2024motor}. Both have well-documented failure modes \citep{dan2025szcore, chavelli2025toward}: sliding windows produce fragmented predictions requiring post-processing, while cascades collapse when the no-event class dominates.} \looseness=-1

\rev{Limited works now formulate event detection as a direct prediction task, deploying one-shot detectors for micro-sleep \citep{dosed}, transformer segmentation for eye-tracking \citep{timeseries_detr}, or sequence-to-sequence models for artefacts and seizures \citep{seeuws2024avoiding}. While these methods demonstrate the value of explicit event parameterization, each remains confined to a limited domain, with no simple way to adapt them for various settings. Therefore, a unified framework for asynchronous EEG decoding is still missing.} \looseness=-1

\rev{To address this problem, we introduce \textbf{\model{}}, an end-to-end framework for detecting and classifying discrete events from raw EEG without relying on pre-aligned windows. \model{} can predict heterogeneous events thanks to a Perceiver-based latent resampling module \citep{jaegle2021perceiver} that compresses each input window into a fixed-size latent grid. This Perceiver is combined with a decoder that performs direct set prediction.}
% Second, hybrid sample/event training objective aligned by a consistency regularizer, where the dense loss supplies the local temporal coherence that bare bipartite matching lacks on EEG. 
\rev{\model{} is evaluated on ten public datasets ($1{,}154$ participants, $2.35\,$M events across six task domains). It achieves a mean event-based F1 score three times higher than the strongest baseline, establishes a new state of the art on seizure monitoring, and matches onset-informed BCI accuracy without using onset information.
This architecture constitutes the first asynchronous neural event identification pipeline that can be easily adapted to heterogeneous settings.
}
\looseness=-1

% Contributions summary
% To solve this problem, we introduce \textbf{\model{}}, an end-to-end framework that adapts the DETR set-prediction paradigm to detect and classify discrete events from raw EEG without pre-aligned windows. 

% \begin{itemize}
%     \item \textbf{Unified architecture:} We integrate a Perceiver-based latent resampling mechanism \citep{jaegle2021perceiver} that allows the same architecture to handle a broad spectrum of event durations, ranging from milliseconds to minutes.
%     \item \textbf{Hybrid training:} We propose a training objective that combines sample-based and event-based prediction, with a consistency regularizer designed to align the two. We demonstrate that this learning objective significantly enhances performance.
%     \item \textbf{Comprehensive set of tasks:} We evaluate the approach on ten independent datasets, encompassing 1{,}154 participants and 2.35 million annotated events across six distinct tasks. \model{} establishes a new state of the art in seizure monitoring and matches specialized time-locked approaches in motor imagery.
% \end{itemize}

% =========================================================================
\section{Event Detection in Time Series}

Formally, we consider a segment of multi-channel EEG recording $\mathbf{X} \in \mathbb{R}^{T \times C}$, where $T$ denotes the number of time samples and $C$ the number of channels. The temporal dimension is defined as $T = f_s \cdot W$, where $f_s$ is the sampling frequency and $W$ the segment duration. 

Unlike window-based methods which assumes a single label for the entire window, our objective is to \textit{detect events}. We aim to map the input segment $\mathbf{X}$ to the set of events it contains, $\mathcal{E} = \{e_i\}_{i=1}^N$. Each event is parameterized as a tuple $e_i = (b_i, c_i)$, where $b_i = (t_{\text{start}}, t_{\text{end}}) \in [0, 1]^2$ denotes the normalized temporal boundaries relative to the segment duration $W$, and $c_i \in \{0, \dots, K\}$ denotes the class of the event, where $K$ is the number of event classes and $c_i=0$ corresponds to the background class. 
Class definitions vary across datasets (see \Cref{tab:datasets_details}).
% The segment duration must strike a balance between coverage and resolution: it should be long enough to include the target events (lower bound), yet short enough to allow accurate boundary regression for each event (upper bound). Additional details are provided in \Cref{app:window_size_constraints}.

There are two ways to evaluate event predictions: using sample-based metrics or using event based metrics.

%The window duration and class definitions vary by dataset, as detailed in \Cref{tab:datasets} (the number of classes includes background) and \Cref{tab:datasets_details}. This formulation facilitates the detection of overlapping events.

\paragraph{Sample-based Metrics}
In many prior studies, the event detection paradigm consists in a timepoint-wise classification task, and the most commonly used metric is the sample-level macro F1 score, called $\mathbf{F1_{\text{sample}}}$ in this work. However, while this formulation aligns with the training objectives of dense predictors (e.g., U-Net), it is unsuitable for quantifying event detection performance. This metric fails to penalize fragmented predictions, meaning that a single event predicted as multiple disjoint segments can yield a high $\mathbf{F1_{\text{sample}}}$ despite failing to capture the event as a coherent unit. We report this metric to enable fair comparisons with prior works.

\paragraph{Event-based Metric}
To assess the model's ability to \textit{decode} neural events, we introduce an event-level F1 score, $\mathbf{F1_{\text{event}}}$, which simultaneously quantifies detection and classification performance.
We define the set of ground-truth events as $\mathcal{E} = \{e_i\}_{i=1}^{N}$ and predicted events as $\mathcal{P} = \{p_j\}_{j=1}^{M}$. A prediction $p_j$ is matched to a ground truth $e_i$ using a sorted greedy matching strategy consistent with standard object-detection protocols \citep{coco}. A match is considered a True Positive if and only if:
\looseness=-1
\begin{equation}
    \text{IoU}(p_j, e_i) > 0.5 \quad \text{and} \quad \text{class}(p_j) = \text{class}(e_i)
\end{equation}
where $\text{IoU}$ denotes the Intersection over Union of the time intervals. Predictions are sorted by class confidence if the IoU with ground-truth exceeds the threshold. A match is accepted only if neither element has already been matched. Unmatched predictions are counted as false positives, and unmatched ground-truth events as false negatives. The final metric is the F1 score derived from these counts.

\rev{$\mathbf{F1_{\text{event}}}$ operates at the event level and is independent of window size. It reflects the number of correctly localized events across the dataset, jointly assessing detection, classification, and temporal precision, and rewarding only accurate matches to ground truth.}
It effectively penalizes fragmented or duplicated predictions.
\rev{In this work, we explicitely target this metric, which better evaluate the method's usability.}
\looseness=-1

% $\mathbf{F1_{\text{event}}}$ is a challenging metric that only rewards precise matches between predictions and ground truth events. It effectively penalizes fragmented or duplicated predictions.

% =========================================================================
\section{Architecture}

We introduce \textbf{\model{}}, an architecture providing an end-to-end alternative to onset-informed EEG decoding by directly framing neural activity as an event-based prediction problem.
As illustrated in \Cref{fig:approach}A, the framework comprises three sequential modules: a convolutional backbone for local feature extraction, a Perceiver-based temporal resampling module for duration adaptation, and a transformer decoder for set-based event prediction.

\paragraph{CNN Backbone: Local Feature Extraction}
The first stage processes the raw EEG input $\mathbf{X} \in \mathbb{R}^{T \times C}$ to extract a sequence of local latent representations for each timestep. For this, we use a modified convolutional architecture based on the work of \citet{defossez}.

The first component is a spatial attention module that is aware of the positions of the sensors, which encodes the $C$ channels to a hidden dimension $D = 270$. It is followed by a sequence of 5 convolutional blocks with a kernel size of 9, which keep the temporal and hidden dimensions unchanged but capture multi-scale temporal dynamics with a dilation factor that increases by a factor of 2 at each layer. Finally, we project the channel dimension from $D$ down to $H = 128$. The output is a sequence of dense embeddings $\mathbf{Z}_{dense} \in \mathbb{R}^{T \times H}$.

\paragraph{Perceiver: Temporal Resampling}

A core challenge in processing heterogeneous neural events is their timescales, spanning from millisecond-level phonemes for speech perception to seizure episodes lasting several minutes. To unify the framework across datasets, we employ a Perceiver module \citep{jaegle2021perceiver}, which translates the sequence of input embeddings of variable length $T$ to a sequence of latents of fixed length $N_{Q_{\text{time}}}$.

The Perceiver maps $\mathbf{Z}_{dense}$ to a latent space of same dimension $\mathbf{Z}_{\text{resamp}} \in \mathbb{R}^{N_{Q_{\text{time}}} \times H}$, via cross-attention over a set of $N_{Q_{\text{time}}}$ learnable latent queries $\mathbf{Q}_{\text{time}} \in \mathbb{R}^{N_{Q_{\text{time}}} \times H}$, effectively acting as a learnable, non-linear downsampling operation. To preserve temporal order, we initialize $\mathbf{Q}_{\text{time}}$ using fixed sinusoidal positional encodings uniformly spaced across the time window.

In our implementation, we set the latent length to $N_{Q_{\text{time}}} = 256$ and use a Perceiver depth of 6, configured with 2 attention heads for both the cross-attention and latent self-attention modules.
Using a smaller $N_{Q_{\text{time}}}$ (e.g., $N_{Q_{\text{time}}} = 128$) makes short events, e.g.\ phonemes, which are on the order of 0.1\,s, impossible to discern from one another, while larger sequence lengths increase memory use quadratically for the downstream decoder. Across datasets, $N_{Q_{\text{time}}} = 256$ strikes the best trade-off between temporal resolution and efficiency.

\rev{Due to the use of a fixed size perceiver, the input segment duration must be adapted for each dataset to strike a balance between coverage and resolution. It should be long enough to include the target events, yet short enough to allow accurate boundary regression for each event. Additional details on how to select the semgent size are provided in \Cref{sec:window_size_constraints}.}

\paragraph{Decoder: Set Prediction}
The final stage decodes the Perceiver's output to generate the final event predictions. We employ a standard transformer decoder as described in \citet{carion2020end} with 4 layers and 4 attention heads. The decoder transforms a set of $N_{Q_{\text{event}}}$ object queries $\mathbf{Q}_{\text{event}}$ into a set of event predictions. Unlike the Perceiver queries, $\mathbf{Q}_{\text{event}}$ are initialized randomly. The inputs can be defined as $\mathbf{Z}_{\text{sparse}} \in \mathbb{R}^{N_{Q_{\text{event}}} \times H}$. After a linear readout, the decoder outputs a set of prediction vectors $\hat{\mathcal{Y}} = \{ \hat{\mathbf{y}}_i \}_{i=1}^{N_{Q_{\text{event}}}}$. Each prediction $\hat{\mathbf{y}}_i = (\hat{b}_i, \hat{c}_i)$ lies in the space $[0, 1]^2 \times \Delta^{K+1}$, concatenating the predicted temporal coordinates $\hat{b}_i$ and the class probability distribution $\hat{c}_i$.

We choose to predict event start and end times rather than center and duration, as this formulation consistently yields better performance. In particular, it naturally mitigates border effects when an event is truncated at the beginning or end of a window: the model can independently predict each boundary, allowing partial events to be handled correctly.

Additional details on preprocessing and training procedures to ensure reproducibility, as well as baseline and ablation models, are provided in \Cref{sec:appendix_methods}. The full model code, along with the training and evaluation pipelines, will be released before the review process.

% =========================================================================
\section{Loss Formulation}
We train the model end-to-end using a loss $\mathcal{L}$ that enforces accuracy both at the dense signal level and the sparse event level, bridged by a consistency regularizer:
\begin{equation}
    \mathcal{L} = \mathcal{L}_{sparse} + \mathcal{L}_{dense} + \lambda_{cons}\mathcal{L}_{cons}
\end{equation}
In our experiments, we balance the contributions of these terms by setting $\lambda_{cons} = 0.5$.

\paragraph{Set Prediction Loss ($\mathcal{L}_{sparse}$)}
To evaluate the set prediction, we first define the pairwise matching cost between a ground truth element $y_i = (b_i, c_i)$ and a prediction $\hat{y}_{\sigma(i)} = (\hat{b}_{\sigma(i)}, \hat{p}_{\sigma(i)})$ indexed by a permutation $\sigma$. Here, $b$ and $c$ represent the normalized temporal segments and class labels, respectively, while $\hat{p}$ is the output of the softmax of the classification head. The matching cost $\mathcal{L}_{match}$ is a linear combination of classification and localization errors. We use the cross-entropy loss $\mathcal{L}_{cls}$ and the IoU loss $\mathcal{L}_{iou}$ to define it:
\begin{equation}
    \mathcal{L}_{match} = \lambda_{cls} \mathcal{L}_{cls}(c_i, \hat{p}_{\sigma(i)}) + \lambda_{iou} \mathcal{L}_{iou}(b_i, \hat{b}_{\sigma(i)})
\end{equation}
We set the balancing coefficients to $\lambda_{cls} = 1$ and $\lambda_{iou} = 5$. \rev{We include a sensitivity analysis of the loss weights in \Cref{fig:loss_weights_sensitivity}.}
The objective is to find a bipartite matching that minimizes the cost between the ground truth set $\mathcal{Y}$ and the predicted set $\hat{\mathcal{Y}}$. Consequently, the final set prediction loss $\mathcal{L}_{event}$ is defined as the minimum over the set of all possible permutations $\mathfrak{S}_{N_{Q_{\text{event}}}}$:
\begin{equation}
    \mathcal{L}_{sparse} = \min_{\sigma \in \mathfrak{S}_{N_{Q_{\text{event}}}}} \sum_{i=1}^{N_{Q_{\text{event}}}} \mathcal{L}_{match}(y_i, \hat{y}_{\sigma(i)})
\end{equation}
This assignment problem is addressed using the Hungarian algorithm \citep{hungarian_matcher}, ensuring that the loss is computed based on the optimal one-to-one correspondence between events and predictions. 
\rev{With this algorithm, the bipartite matching problem can be solved in $\mathcal{O}\left(N_{Q_{\text{event}}}^3\right)$.}

\paragraph{Dense Auxiliary Loss ($\mathcal{L}_{dense}$)}
Dense supervision methods are typically easier to optimize and yield meaningful predictions earlier in the training process. To combine the strengths of dense and sparse paradigms, we additionally attach a linear probe to the output of the Perceiver $\mathbf{Z}_{resamp}$ to predict the class for every timestep, giving an auxiliary cross-entropy loss:
\begin{equation}
    \mathcal{L}_{dense} = \frac{1}{N_{Q_{\text{time}}}} \sum_{t=1}^{N_{Q_{\text{time}}}} \mathcal{L}_{ce}(\hat{y}_{t, dense}, y_{t, true})
\end{equation}

\paragraph{Consistency Loss ($\mathcal{L}_{cons}$)}
\label{sec:consistency_loss}
To enforce coherence between sample-based and event-based predictions, we project the discrete event-level predictions from the Decoder onto the temporal axis to create a reconstructed dense probability map $P_{\text{event}}$. For each predicted event, its predicted class probability $\hat{p}$ is assigned uniformly onto the temporal interval defined by its predicted boundaries $\hat{b}$. When multiple events overlap in time, their class probabilities are summed and normalized at each timestep to form a valid probability distribution. The consistency loss is defined as the Kullback--Leibler (KL) divergence between the reconstructed event distribution $P_{\text{event}}$ and the timestep-level probability distribution $P_{\text{dense}}$:
\looseness=-1
\begin{equation}
\mathcal{L}_{cons} = D_{KL} \left( P_{\text{dense}} \parallel P_{\text{event}} \right)
\end{equation}

This auxiliary objective encourages event-level outputs to be coherent with the dense predictions. The complete projection procedure and additional details are provided in \Cref{sec:appendix_consistency}.

% =========================================================================
\begin{table}[h]
\caption{Summary of experimental datasets. Mean values are displayed.}
\label{tab:datasets}
\centering
\small
\renewcommand{\arraystretch}{1.2}
\setlength{\tabcolsep}{6pt}
\begin{tabular}{llcc}
\toprule
\textbf{Task Domain} & \textbf{Dataset} & \textbf{\# Events / Window} & \textbf{Event Duration (s)} \\
\midrule
\multirow{2}{*}{\textbf{Speech}}
 & Broderick \citep{broderick} & 75.33 (Phonemes) & 0.08 \\
 & SparrKULee \citep{accou} & 19.5 (Words)     & 0.3 \\
\midrule
\textbf{Typing} & Pinet \citep{lucy_typing} & 19.3 & 0.11 \\
\midrule
\multirow{3}{*}{\textbf{P300}}
 & BI2013   \citep{congedo2011} & 60.1 & 1 \\
 & BI2014a  \citep{korczowski_a} & 45.8 & 1 \\
 & BI2014b  \citep{korczowski_b} & 9.9  & 1 \\
\midrule
\multirow{2}{*}{\textbf{Motor Imagery}}
 & BNCI2014 \citep{tangermann2012}  & 1.3  & 4 \\
 & BNCI2015 \citep{faller2012} & 3.4 & 5 \\
\midrule
\textbf{Artifacts} & TUAR (3.0.1) \citep{hamid2020} & 11.03 & 9.04  \\
\midrule
\textbf{Seizure} & TUSZ (2.0.3) \citep{shah2018} & 0.25 & 67.7 \\
\bottomrule
\end{tabular}
\end{table}

\begin{figure}[t]
    \centering
    \includegraphics[width=\textwidth]{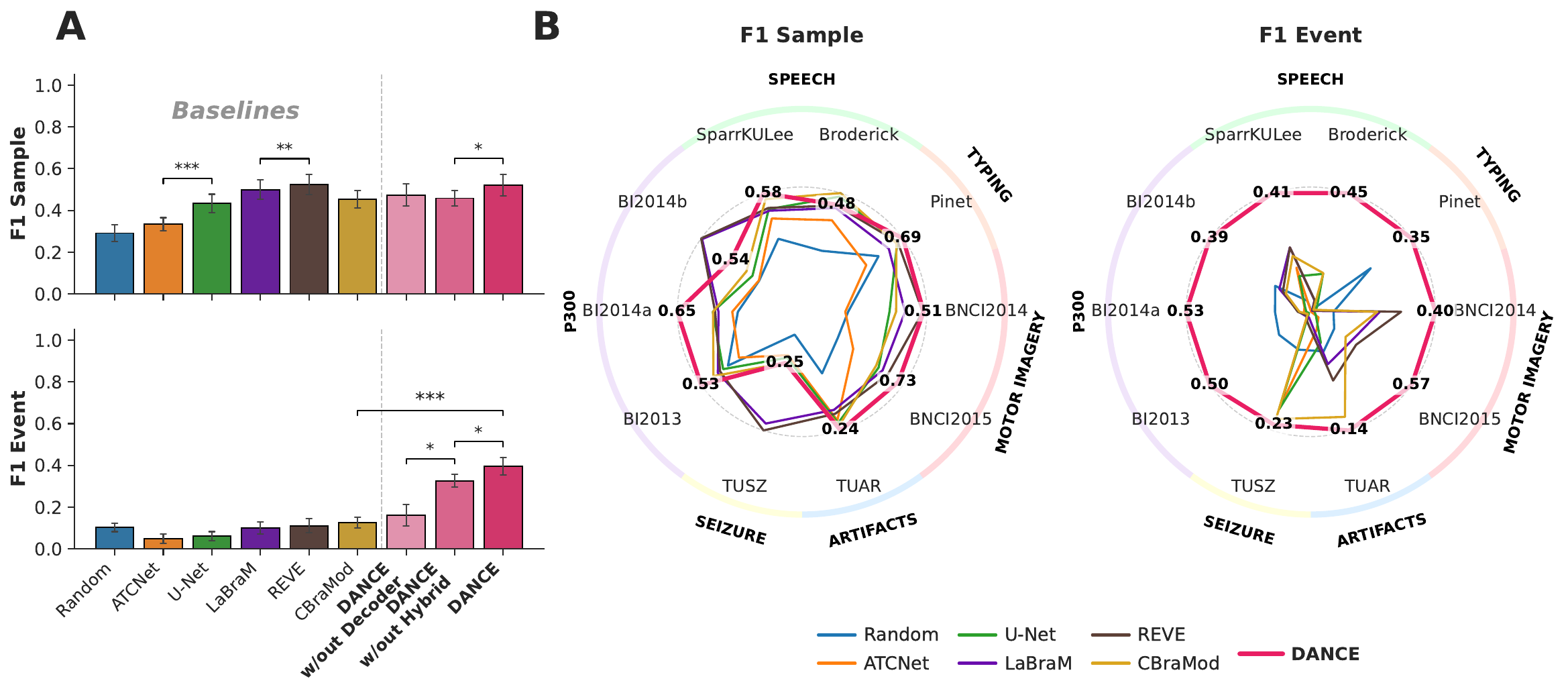}
    \caption{
    \textbf{\rev{\model{} yields a threefold improvement in event-based F1 over all baselines.}}\\
    \textbf{A.} Performance averaged across subjects and datasets (Mean $\pm$ SEM) comparing baselines and ablations against the proposed \model{} model. Statistical significance via Linear Mixed-Effects Models (LMM; \Cref{sec:statistics}) is indicated as follows: ${}^{*}p < 0.05$, ${}^{**}p < 0.01$, and ${}^{***}p < 0.001$.
    \textbf{B.} Normalized radar profiles showing relative averaged performance per subject across individual datasets. Axes are normalized to the best-performing method for each dataset. Numeric annotations indicate the absolute score of \model{} at each vertex, and the outer ring groups datasets by task category.
    }
    \label{fig:performance}
\end{figure}

\section{Results}
To validate our model, we evaluate its performance on 10 datasets, described in \Cref{fig:approach}C in terms of data volume and in \Cref{tab:datasets} for key characteristics. To probe performance on fine-grained, high-frequency events, we include Speech (Broderick, SparrKULee) and Typing (Pinet) tasks, in which event durations are very short. We also include BCI datasets, with events ranging from 1 to 5 seconds. For the P300 paradigm from the Brain Invaders suite (BI2013, BI2014a, BI2014b), the number of events per window is high and we detect both target and non-target visual stimuli. For Motor Imagery datasets (BNCI2014, BNCI2015), events are scarcer and provide longer-duration events associated with endogenous cognitive processes. To assess \model{} capabilities on longer events, we include the Temple University Artifact corpus (TUAR) with overlapping long artifacts, and the Temple University Seizure (TUSZ) dataset---the largest open-source corpus for seizure monitoring---where precise end-to-end event identification is of critical diagnostic value.

\paragraph{Preprocessing}
To ensure consistency across domains, we apply a unified preprocessing pipeline to all datasets, following the minimal protocol proposed by \citet{levy2025}. Signals are first bandpass-filtered between $0.1$ and $100$\,Hz then resampled to a common rate of $f_s = 128$\,Hz. Subsequently, we apply robust normalization to each channel within each session. This is achieved using a \texttt{RobustScaler} \citep{pedregosa2011scikit}, which subtracts the median and divides by the interquartile range, followed by a clamping operation to restrict values to the interval $[-16, 16]$.

\paragraph{Evaluation methods}
We employ a cross-subject splitting for all datasets. More implementation details for model training and evaluation can be found respectively in \Cref{sec:general_appendix_details,sec:appendix_evaluation}.

We first compare the overall performance of \model{} to \rev{six} baseline approaches. First, a Random model informed by dataset statistics, which randomly samples $k$ events from the total event pool, where $k$ is the window duration multiplied by the mean event frequency. This is a strong baseline as it leverages real event distributions. Second, a modified window-based model: ATCNet \citep{altaheri2022physics}. Third, a segmentation-based model: USleep \citep{perslev2021u}. Finally, \rev{three} EEG foundation models: \rev{LaBraM \citep{labram}}, CBraMod \citep{wang2024cbramod} and \rev{REVE \citep{elouahidi2025reve}}. We also compare \model{} against two ablations: \model{} w/out Decoder, to assess the benefit of the sparse view, and \model{} w/out Hybrid to evaluate the impact of our hybrid learning objective. 
\rev{A challenge for fair comparison is that most available models only report onset-informed performances. For both window-based and foundation models, we adapted them to the continuous setting by generating timestep-level predictions, in particular replacing the final classifier with a dense prediction layer for foundation models.} More information on baselines, ablations, and tuning can be found in \Cref{sec:appendix_gap_filling,sec:appendix_baselines,sec:appendix_ablations}.

The evaluation reports two metrics: $\mathbf{F1_{\text{event}}}$ and $\mathbf{F1_{\text{sample}}}$ over the 10 datasets. \Cref{fig:performance} summarizes these results, presenting both the aggregated performance distributions across datasets (A) and detailed per-dataset outcomes (B).
% DANCE versus Baselines
\paragraph{Sample-based and event-based performance}
% Our model significantly outperforms the strongest baseline, CBraMod, on both metrics. \model{} achieves a mean $\mathbf{F1_{\text{sample}}}$ of $0.521 \pm 0.052$, a 15\% improvement over CBraMod. While this gap is significant ($p=0.003$ with Linear Mixed-Effects Models), the event-based metric is where \model{} stands out, achieving $\mathbf{F1_{\text{event}}} = 0.397 \pm 0.042$, a threefold improvement over CBraMod. This result underscores the importance of an event-based objective for detection.
\rev{\model{} significantly outperforms every baseline on the event-based metric, reaching $\mathbf{F1_{\text{event}}} = 0.397 \pm 0.042$ --- a threefold improvement over the strongest baseline, CBraMod ($0.126$; $p < 10^{-6}$ with Linear Mixed-Effects Models). On the sample-based metric, results are closer: \model{} reaches $\mathbf{F1_{\text{sample}}} = 0.521 \pm 0.052$, statistically tied with REVE and LaBraM and significantly above CBraMod ($p = 0.006$).} The detailed performance breakdown is summarized via radar plots normalized by the highest performance (\Cref{fig:performance}B). The results show that a model can achieve strong performance on sample-based metrics while performing poorly on event-based metrics. Notably, \model{} avoids the catastrophic failure modes observed in traditional attention or CNN approaches (e.g., ATCNet on BNCI2014 or U-Net on BI2014b).

% Real-time metrics
\paragraph{Distance-based performance}
Additionally, we report a class-agnostic detection F1 ($F1_{\mathrm{det}}$) and a median center-point error ($\Delta_{\mathrm{ctr}}$, in seconds) to provide an interpretation of model performance in physical time units. (\Cref{fig:metric_robustness}B and \Cref{tab:reviewer_metrics}). \model{} dominates the best baseline on \emph{both} axes on all $10$ datasets, with median center localization staying within $25\%$ of event duration in every case (e.g.\ $14$\,ms on $80$\,ms phonemes, $78$\,ms on $4$-second motor-imagery trials, and $4.5$\,s on multi-minute seizures). The best baseline meanwhile is $2$ to $19{\times}$ less precise than \model{} on cognitive tasks and $3{\times}$ on BCI motor imagery. The recall gap is very large on long events: \model{} retrieves $10{\times}$ more seizures on TUSZ than the best baseline.

% DANCE versus Ablations
\paragraph{Ablation experiments} 
Finally, the ablation results show the benefit of our architectural choices. We observe a clear, statistically significant improvement for each block on the $\mathbf{F1_{\text{event}}}$ metric. The base \model{} w/out Decoder model ($0.161$), using only dense predictions, is significantly outperformed by the \model{} w/out Hybrid variant ($0.326$, $p=0.015$), confirming the critical role of the set-based prediction of events compared to dense approaches. Furthermore, the full \model{} model with its hybrid optimization also improves performance compared to the decoder-only configuration ($0.398$, $p=0.025$), confirming that the modified learning objective is essential for optimal performance.

\begin{figure}[h]
    \centering
    \includegraphics[width=1.0\linewidth]{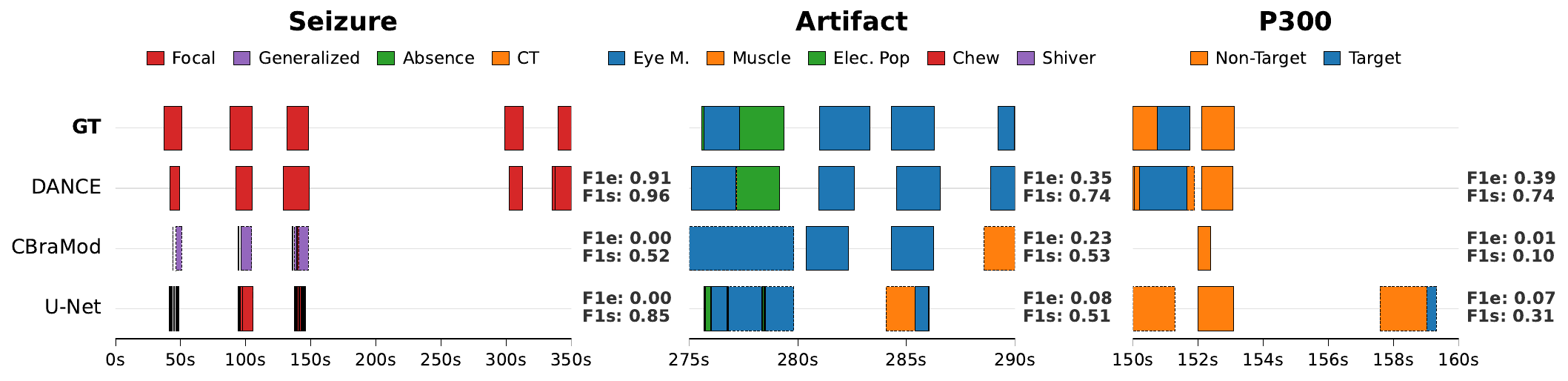}
    \caption{
        \textbf{\model{} yields qualitatively superior event predictions than baseline architectures.}
        Visualization of selected session segments comparing three architectures: U-Net, CBraMod, and our proposed model. Ground truth (GT) events are depicted alongside predictions from the respective models. Solid outlines indicate true positives, while dashed outlines denote false positives with respect to $\mathbf{F1_{\text{event}}}$. Colors represent distinct classes. $\mathbf{F1_{\text{event}}}$ (F1e) and $\mathbf{F1_{\text{sample}}}$ (F1s) are shown on the right for each portion.
    }
    \label{fig:decoded_examples}
\end{figure}
\paragraph{Visualization of decoded events}
\Cref{fig:decoded_examples} illustrates the temporal decoding capabilities of our model compared to the CBraMod and U-Net baselines. We selected three datasets (TUSZ, TUAR, BI2014b) to demonstrate the versatility of our architecture across diverse signal characteristics, protocols, and event durations.

The \rev{left} panel highlights performance on long-duration events, showing a 350-second window containing focal seizures (red). Our model successfully detects and classifies all five seizure events. Conversely, the baselines exhibit significant degradation; they detect the initial three events but fail to classify them correctly and completely miss the subsequent two seizures. Both CBraMod and U-Net yield fragmented predictions. In particular, $\mathbf{F1_{\text{sample}}}$ fails to penalize U-Net while $\mathbf{F1_{\text{event}}}$ does. It is worth noting that even with heuristic post-processing (e.g., gap-filling with varying thresholds), these fragmented outputs fail to improve the $\mathbf{F1_{\text{event}}}$ metric. We detail the impact of post-processing methods in \Cref{sec:appendix_gap_filling}.

The middle panel shows a 15-second segment from the TUAR dataset. \model{} produces the most precise predictions, correctly distinguishing between different eye-movements and identifying an electrode-pop artifact. In contrast, the two baseline models fail to separate the artifacts types and both detect a muscle artifact that is not present in the signal.

The \rev{right} panel examines one of the Brain Invaders P300 dataset (BI2014b). This task involves distinguishing between non-target (orange) and target (blue) events lasting one second. Our model identifies the non-target events and provides an approximate localization of the target event. CBraMod fails to detect these events entirely. U-Net offers reasonable localization but struggles with class specificity, generating a false positive around the 158\,s mark.

These visualizations highlight the model's ability to generate temporally coherent event proposals across a wide range of event durations, inter-event gaps, and signal complexities, without relying on additional post-processing. \rev{We show other examples on BCI datasets in \Cref{fig:decoded_examples_bnci}}.

\paragraph{Seizure Monitoring}
\begin{figure}[h]
    \centering
    \includegraphics[width=1.0\linewidth]{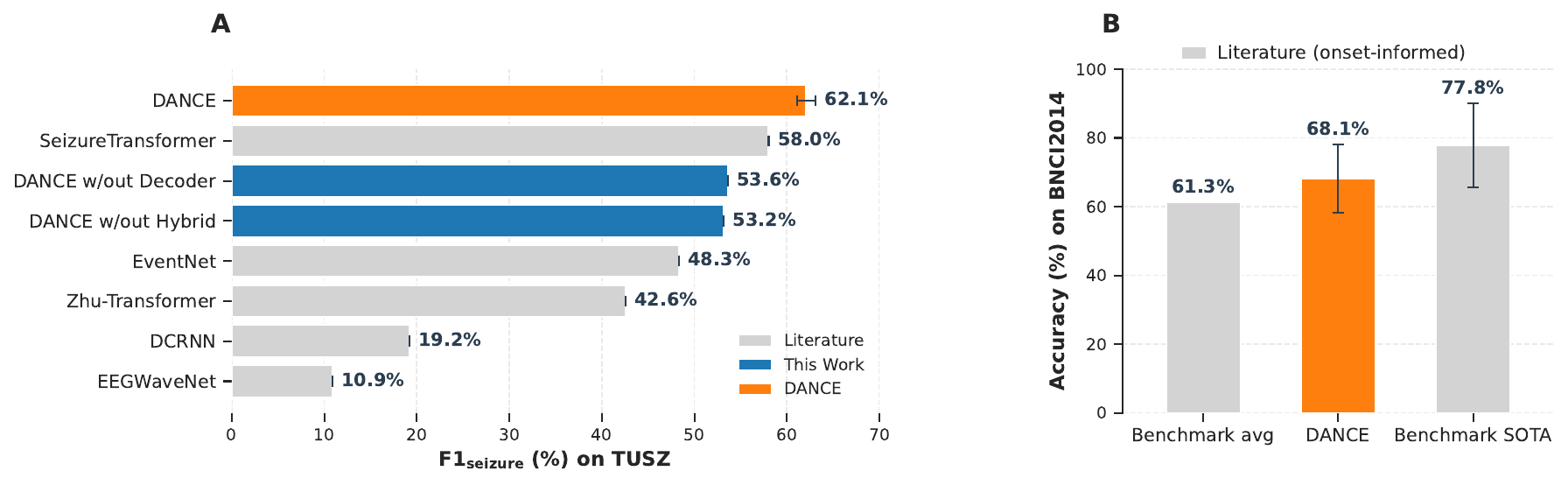}
    %\vskip-1em
    \caption{
    \rev{\textbf{\model{} establishes a new state-of-the-art for clinical seizure monitoring on the TUSZ dataset and matches onset-informed approaches on BNCI2014.} 
    \textbf{A.} Benchmarking of \model{} against existing literature and ablation baselines using the $\mathbf{F1_{\text{seizure}}}$ metric. Scores are reported from \citet{wu2025seizuretransformer}. \model{} is evaluated across 3 model seeds. TUSZ has a fixed test split. \textbf{B.} Intra-session performance on the BNCI2014 dataset for \model{} and the MOABB \citep{moabb} benchmark.} Standard deviation is reported.
    }
    \label{fig:seizure}
\end{figure}

We further validate our model in the context of seizure monitoring. We benchmark our approach against recent methods for event-based segmentation of seizures: SeizureTransformer \citep{wu2025seizuretransformer}, EventNet \citep{seeuws2024avoiding}, ZhuTransformer \citep{zhu_transformer}, DCRNN \citep{dcrnn}, and EEGWaveNet \citep{eegwavenet}. We report the mean performance across three runs with different seeds on the fixed test set in \Cref{fig:seizure}A. Our model outperforms these prior works on the $\mathbf{F1_{\text{seizure}}}$\footnote{For the seizure detection task, we use the binary F1 score $\mathbf{F1_{\text{seizure}}}$ to avoid over-reliance on the dominant background class, in line with recent recommendations \citep{dan2025szcore}.} score, reaching 62.1\% on the fixed test set vs.\ 58.0\% for the second-best method. By increasing the confidence threshold to strongly suppress false alarms (threshold = 0.99), DANCE achieves 1.88 false positives per 24 hours—a key clinical metric—while maintaining an $\mathbf{F1_{\text{seizure}}}$ above 58.0\%, highlighting its potential clinical relevance.

\paragraph{\model{} versus Onset-informed Approaches}

% \begin{table}[H]
%     \centering
%     \caption{Intra-session performance on the BNCI2014 dataset.}\vspace{0.5em}
%     \label{tab:intra_session_perf}
%     \setlength{\tabcolsep}{10pt}
%     \footnotesize
%     \begin{tabular}{cccc}
%         \toprule
%         \textbf{Method} & \textbf{F1 score} & \textbf{Accuracy} \\
%         \midrule
%         Benchmark avg  & --      & 61.1\% \\
%         \model{}       & 96.57\% & 68.14\% \\
%         Benchmark SOTA & --      & 77.8\% \\
%         \bottomrule
%     \end{tabular}
% \end{table}

How does our end-to-end approach compare to onset-informed pipelines? As established in the comprehensive benchmark by \citet{moabb}, state-of-the-art methods on BNCI2014 achieve an average classification accuracy of 61.3\% with intra-session splitting, with the best-performing model scoring 77.82\%. We use the same protocol for comparison. We define accuracy on match as the classification accuracy computed only over predictions that are temporally matched to a ground-truth event. Our model, which simultaneously detects and classifies, achieves an accuracy on match of 68.14\%, supported by a standard F1 score on all predictions of 96.57\% (Recall = 98.55\%, Precision = 94.69\%; \Cref{fig:seizure}B). This result suggests that, on highly structured datasets like BNCI2014, competitive decoding performance can be achieved without onset information.
\looseness=-1

\begin{figure}[!t]
    \centering
    \includegraphics[width=\linewidth]{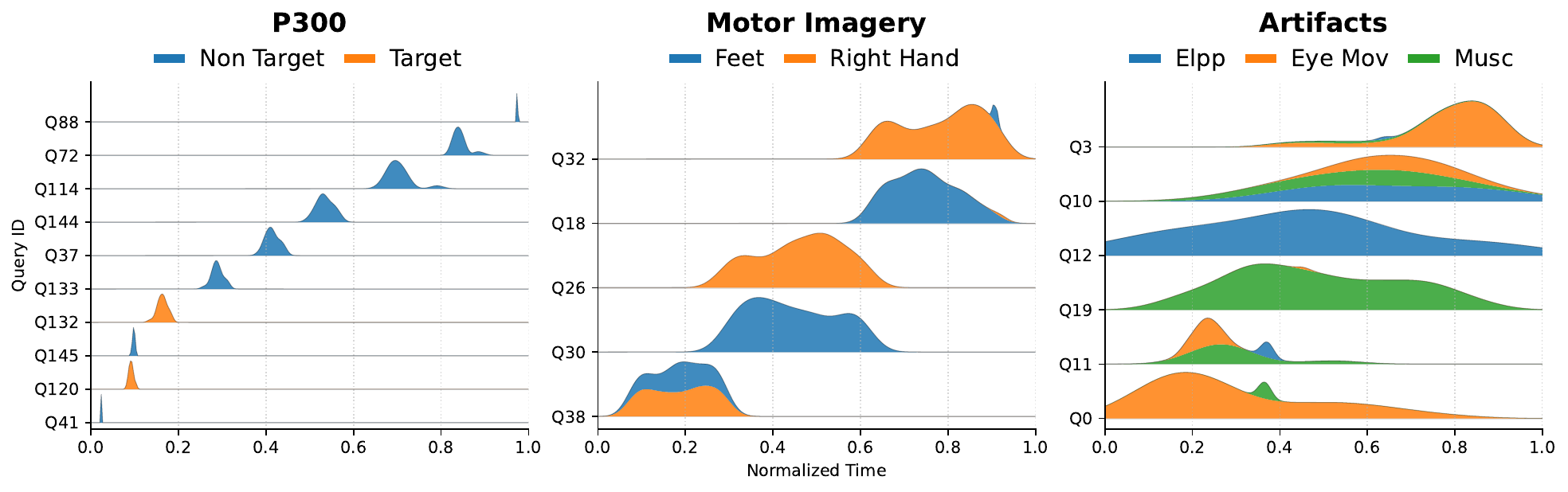}
    \caption{
    \textbf{\model{} queries dynamically specialize in specific temporal regions and event classes.}
    Visualization of the temporal distribution of predictions across selected queries. \rev{For each query row, the outer envelope corresponds to when the query activates. This is the total density of predictions. The colored bands indicate what it predicts. The stacked inner colored bands decompose this density by predicted class (most frequent at the base). A sharp, narrow envelope indicates temporal specialization, while a dominant color indicates class specialization.}
    }
    \label{fig:query_specialization}
\end{figure}

\paragraph{Query Specialization}

By inspecting the learned queries, we can gain insights into how \model{} specializes across event classes and temporal localization. \Cref{fig:query_specialization} shows selected event queries across three datasets, where we visualize the distribution of predicted event centers: peaked distributions indicate temporal specialization, with queries consistently attending to narrow regions of the time window, while color composition reflects class specialization, with monochromatic distributions indicating queries that exclusively detect a specific class. The results for the P300 task (BI2013) are displayed on the left, with 10 of the 150 queries shown. We observe both temporal and class specialization, consistent with findings in computer vision models \citep{carion2020end}. Among queries sharing similar temporal localizations (e.g., Q145 and Q120), we observe specialization to different classes, indicating that queries effectively disentangle detection and classification. The middle panel shows Motor Imagery (BNCI2015), where temporal specialization remains clear but with broader distributions. Some queries are class-specific (Q30, Q18), while others attend to both classes (Q38). Finally, the right panel examines the Artifacts task: while some queries remain class-specialized (Q19, Q3), temporal distributions are more relaxed, with broad overlapping receptive fields rather than sharp localization, and some queries exhibit hybrid behavior by attending to multiple artifact types (Q10). Overall, these findings suggest that \model{} does not enforce a rigid template, but instead adapts its querying mechanism to the structure of each dataset.
\looseness=-1

% =========================================================================
\section{Discussion}
\label{sec:discussion}

% Result
In this work, we introduce \model{}, an end-to-end model to jointly detect and classify events from neural time series.
Thanks to its flexible architecture, \model{} can be readily applied to a variety of tasks and event types. \rev{The fact that DANCE achieves good performance without onsets and with a common architecture is an improvement over the current practices in EEG where specialized onset-based architectures are used for each task (e.g., SeizureTransformer \citep{wu2025seizuretransformer} for seizure or ATCNet \citep{altaheri2022physics} for BCI.)}
%We validate its performance on a comprehensive benchmark, encompassing a variety of clinical, cognitive and BCI tasks. 
By framing event detection as a set prediction problem, rather than as a timestep-wise classification task \rev{and adapting it to the noisy and heterogeneous EEG setting}, we shift towards more realistic settings and show that time-locked scenarios are not a prerequisite for high-performance decoding. 

%where ground-truth timings are difficult or impossible to obtain, often requiring adaptations of the experimental protocol. For example, imagination experiments often require cues to mark the beginning or end of an event, either imposed or triggered by the participant. This not only introduces confounds in the experimental setting, but also limits applicability to realistic use cases.

% Limits
\paragraph{Limitations.}
While our model demonstrates capabilities in identifying temporal boundaries and classes of diverse events, specific challenges remain. \rev{First, for very short events, such as phonemes or keystrokes, we restrict the task to voice detection.
Datasets with very short events are characterized by a high event density. In this context with low-SNR EEG, even onset-based approaches struggle to correctly classify events without including broader context \citep{defossez}, language priors \citep{zhang2025decoding} or higher SNR modalities \citep{emg2qwerty,willett_handwriting}.}
% For such fine-grained events, high-precision decoding may require onset information, which can guarantee perfect phase-locking, or higher SNR modalities such as MEG \citep{defossez, levy2025}, intracranial recordings \citep{willett_handwriting}, or EMG \citep{emg2qwerty}.
Second, while our unified architecture can be adapted to a variety of tasks without any changes, we have not yet observed any improvements from jointly training across datasets, likely due to the heterogeneity and limited size of the datasets considered. \rev{Addressing this limitation constitutes a natural next research direction.} However, our method paves the way towards large-scale pretraining for event detection --- a task for which current EEG foundational models fall short, \rev{as evidenced by CBraMod \citep{wang2024cbramod}, REVE \citep{elouahidi2025reve} and LaBraM \citep{labram} which substantially underperform our pipeline.}

% % Conclusion and Future Steps
% \paragraph{Conclusion.}
% This work represents a step towards flexible BCI systems capable of decoding continuous streams without rigid experimental constraints. We aim to build upon these findings to advance cognitive decoding, enable real-time applications, and propose a seamless pipeline for brain decoding.

% =========================================================================
% NeurIPS expects "Broader Impacts" rather than the ICML "Impact Statement";
% the content from the ICML submission is preserved.

% =========================================================================
\newpage
% arXiv-ready: read the pre-built .bbl directly so dance_neurips.bib is
% not needed on the arXiv server and bibtex is never invoked there.
% To refresh dance_meta.bbl after editing dance_neurips.bib, temporarily
% replace the \input below with the two commented lines, run `make`,
% then revert.

% \bibliographystyle{plainnat}
% \bibliography{dance_neurips}

% =========================================================================
%                                APPENDIX
% =========================================================================
\newpage
\appendix
\counterwithin{figure}{section}
\def\thefigure{\thesection.\arabic{figure}}
\counterwithin{table}{section}
\def\thetable{\thesection.\arabic{table}}

\section{Complementary Analysis}
\label{sec:comp_analysis}

\subsection{Metric robustness and decoupled analysis}
\label{sec:metric_robustness}

\rev{$F1_{\mathrm{event}}$ groups three facets of a prediction: detection, classification, and temporal precision. We decouple it with two complementary physical-unit metrics:
\begin{itemize}
    \item $F1_{\mathrm{det}}$: class-agnostic detection at
    $\mathrm{IoU} \geq 0.5$, isolating detection from classification.
    \item $\Delta_{\mathrm{ctr}}$: median center-point error of
    matched true positives, in seconds, isolating temporal precision
    from detection and classification.
\end{itemize}}

\begin{figure}[h]
    \centering
    \includegraphics[width=1.0\linewidth]{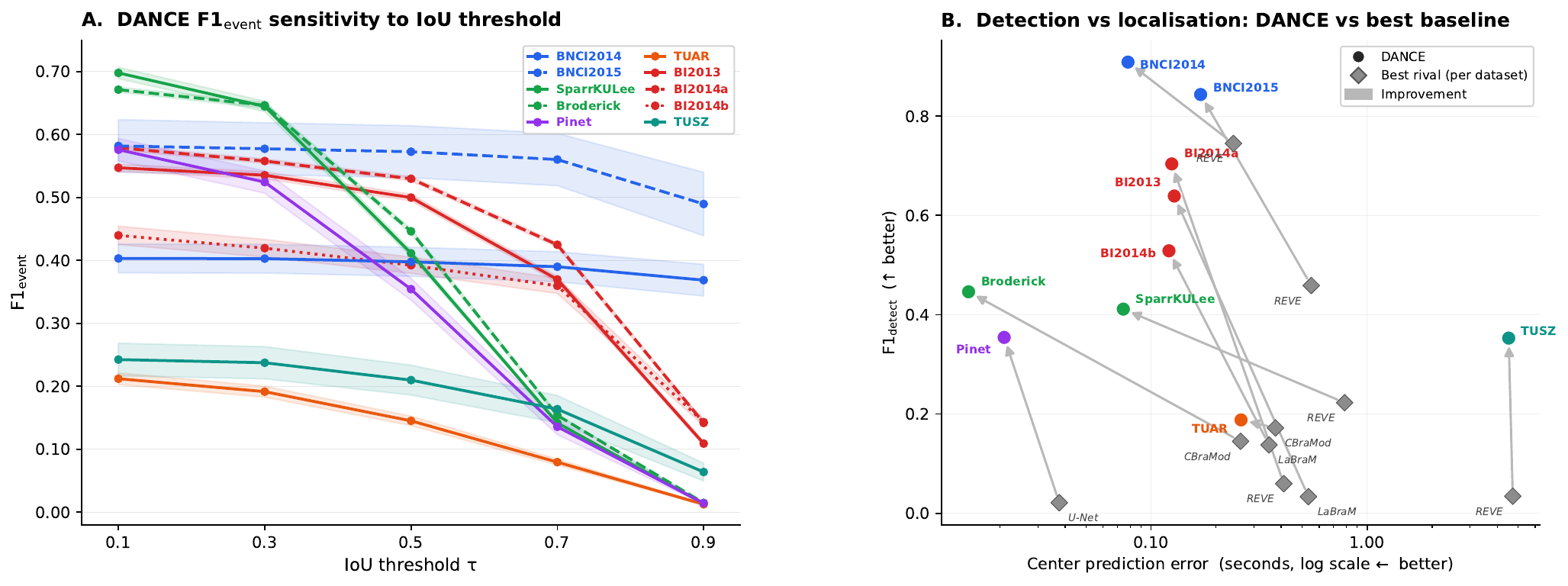}
    \caption{\rev{\textbf{A.} $F1_{\mathrm{event}}$ of \model{} as a
    function of the IoU acceptance threshold $\tau$, one curve per dataset (coloured by task domain). \textbf{B.} Detection quality
    ($F1_{\mathrm{det}}$, class-agnostic, $\uparrow$ better) versus
    temporal precision ($\Delta_{\mathrm{ctr}}$, center-point error in seconds, log scale, $\leftarrow$ better) for \model{} (coloured) against the best non-ablation rival (grey). If the arrow points to the upper left, it means DANCE dominates on both axes. The arrow runs from the best baseline to the DANCE model.}}
    \label{fig:metric_robustness}
\end{figure}

\rev{Panel A shows that $F1_{\mathrm{event}}$ decays monotonically and
smoothly with $\tau$ on all ten datasets, with no cliff between the
$\tau = 0.5$ default and the stricter $\tau = 0.9$ regime: the chosen
operating point is not a knife-edge.} 

\rev{Panel B and \Cref{tab:reviewer_metrics} report the physical-unit picture per dataset: \model{} dominates the best rival on both axes on \emph{every one of the 10 datasets}. In absolute terms, $\Delta_{\mathrm{ctr}}$ is $14$--$74$\,ms on speech (Broderick, SparrKULee), $21$\,ms on typing (Pinet), $\sim 125$\,ms across the three P300 datasets, $78$--$170$\,ms on BCI motor-imagery, $261$\,ms on TUAR artifacts, and $4.5$\,s on TUSZ seizures. Relative to event duration, this amounts to $17$--$25\%$ on sub-second speech and typing events, $\sim 12\%$ on P300 stimuli, and $\leq 7\%$ on every other dataset. This is consistent with the main reported results. The predicted event center sits well inside the true event's temporal support, which is the precision required to time-lock a prediction to the underlying neural response for downstream analysis (ERP averaging, BCI feedback, clinical onset tagging). $F1_{\mathrm{det}}$ is meanwhile $3{\times}$ to $19{\times}$ higher than the best rival on $6/10$ datasets (e.g.\ BI2013: $0.64$ vs.\ $0.03$ for LaBraM), and exceeds it on every dataset, so this precision is achieved while retrieving strictly more events.}

\begin{table}[H]
\centering
\caption{Decoupled detection and localization metrics. F1\textsubscript{detect} measures class-agnostic detection (IoU$\geq$0.5). F1\textsubscript{event} adds class awareness. $\Delta$center is the mean temporal error of true positives in seconds. Best rival is the highest-scoring non-ablation baseline per dataset.}
\label{tab:reviewer_metrics}
\small
\setlength{\tabcolsep}{4pt}
\begin{tabular}{l l  c c c  c c c}
\toprule
& & \multicolumn{3}{c}{\textbf{DANCE}} & \multicolumn{3}{c}{\textbf{Best Rival}} \\
\cmidrule(lr){3-5} \cmidrule(lr){6-8}
Dataset & Rival & F1\textsubscript{det} & F1\textsubscript{evt} & $\Delta$ctr (s) & F1\textsubscript{det} & F1\textsubscript{evt} & $\Delta$ctr (s) \\
\midrule
BNCI2014 & REVE & 0.909 & 0.398 & 0.078 & 0.745 & 0.288 & 0.241 \\
BNCI2015 & REVE & 0.843 & 0.573 & 0.170 & 0.459 & 0.257 & 0.551 \\
\addlinespace[3pt]
SparrKULee & REVE & 0.411 & 0.411 & 0.074 & 0.223 & 0.223 & 0.787 \\
Broderick & CBraMod & 0.446 & 0.446 & 0.014 & 0.145 & 0.145 & 0.260 \\
\addlinespace[3pt]
Pinet & U-Net & 0.354 & 0.354 & 0.021 & 0.021 & 0.021 & 0.038 \\
\addlinespace[3pt]
TUAR & CBraMod & 0.188 & 0.145 & 0.261 & 0.172 & 0.122 & 0.377 \\
\addlinespace[3pt]
BI2013 & LaBraM & 0.639 & 0.500 & 0.128 & 0.034 & 0.027 & 0.535 \\
BI2014a & REVE & 0.704 & 0.530 & 0.125 & 0.060 & 0.055 & 0.411 \\
BI2014b & LaBraM & 0.529 & 0.392 & 0.121 & 0.138 & 0.122 & 0.352 \\
\addlinespace[3pt]
TUSZ & REVE & 0.353 & 0.210 & 4.522 & 0.034 & 0.020 & 4.720 \\
\bottomrule
\end{tabular}
\end{table}

\subsection{Sensitivity to the DETR loss weights}
\label{app:loss_weights_sensitivity}

\begin{figure}[h]
    \centering
    \includegraphics[width=0.75\linewidth]{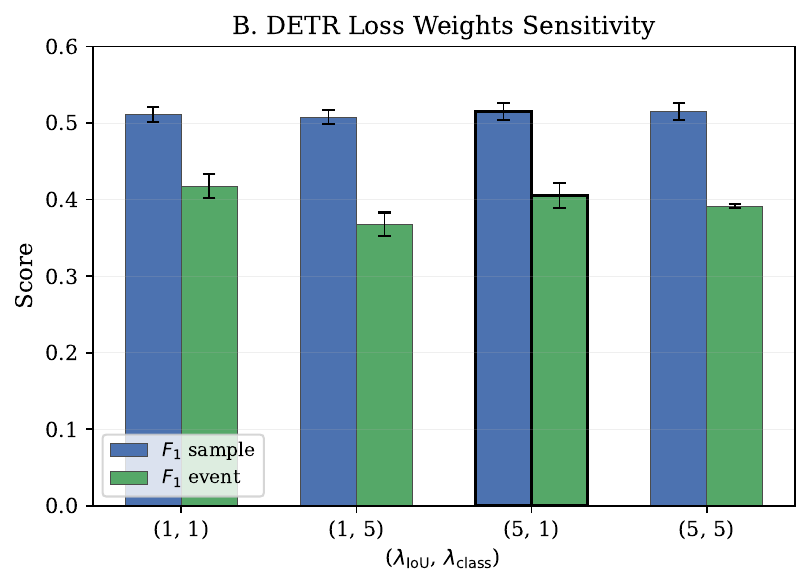}
    \caption{\textbf{\rev{DANCE is robust to the DETR loss weights.}}
    \rev{Mean $F1_{\mathrm{sample}}$ (blue) and $F1_{\mathrm{event}}$
    (green) over BI2014b, BNCI2014 and Broderick (5-fold CV per
    dataset) for four configurations of the DETR localization and
    classification loss weights
    $(\lambda_{\mathrm{IoU}}, \lambda_{\mathrm{class}})$.
    Error bars are the SEM across datasets. The default configuration
    $(5, 1)$ used in every other experiment is highlighted with a thick
    border. Performance varies by less than $0.05$ F1 across the swept
    range.}}
    \label{fig:loss_weights_sensitivity}
\end{figure}

\rev{The DETR component of the hybrid loss balances temporal localization ($\lambda_{\mathrm{IoU}}$) and event classification ($\lambda_{\mathrm{class}}$).
We selected the default configuration $(\lambda_{\mathrm{IoU}}, \lambda_{\mathrm{class}})
= (5, 1)$ to bias optimisation toward precise temporal boundaries before refining the discrete classification of those bounded segments. To show
that the model is robust to this choice, we re-trained DANCE on the four corners of the $(\lambda_{\mathrm{IoU}}, \lambda_{\mathrm{class}}) \in \{1, 5\}^2$ grid on three datasets (BI2014b, BNCI2014, Broderick), keeping every other hyperparameter fixed and using the same 5-fold cross-validation protocol as in the main experiments. Figure~\ref{fig:loss_weights_sensitivity} reports both $F1_{\mathrm{sample}}$ and $F1_{\mathrm{event}}$ averaged across the three datasets. The two metrics vary by less than $0.05$ F1 across the swept range, confirming the robustness}

\subsection{Window size is determined by architectural constraints}
\label{sec:window_size_constraints}

\rev{The choice of the window size is fully determined by two architectural constraints:
\begin{itemize}
    \item \textbf{Coverage (lower bound):} $W$ must be long enough to
    fully contain the target events. For example, the $256$\,s window
    used on TUSZ is required because the average seizure duration is
    $67.7$\,s and routinely exceeds four minutes.
    \item \textbf{Resolution (upper bound):} \model{} resamples every
    window to a fixed latent of $N_{Q_{\text{time}}} = 256$ tokens
    (\Cref{sec:general_appendix_details}). Accurate boundary regression
    requires each event to be represented by at least three latent
    tokens, yielding the rule of thumb
    \[
        \frac{N_{Q_{\text{time}}}}{\bar{n}}
        \;=\; \frac{256}{\bar{n}} \;\geq\; 3\,,
    \]
    where $\bar{n}$ is the mean number of events per window.
\end{itemize}}

\begin{figure}[h]
    \centering
    \includegraphics[trim=0cm 0cm 0cm 0.67cm,clip,width=0.7\linewidth]{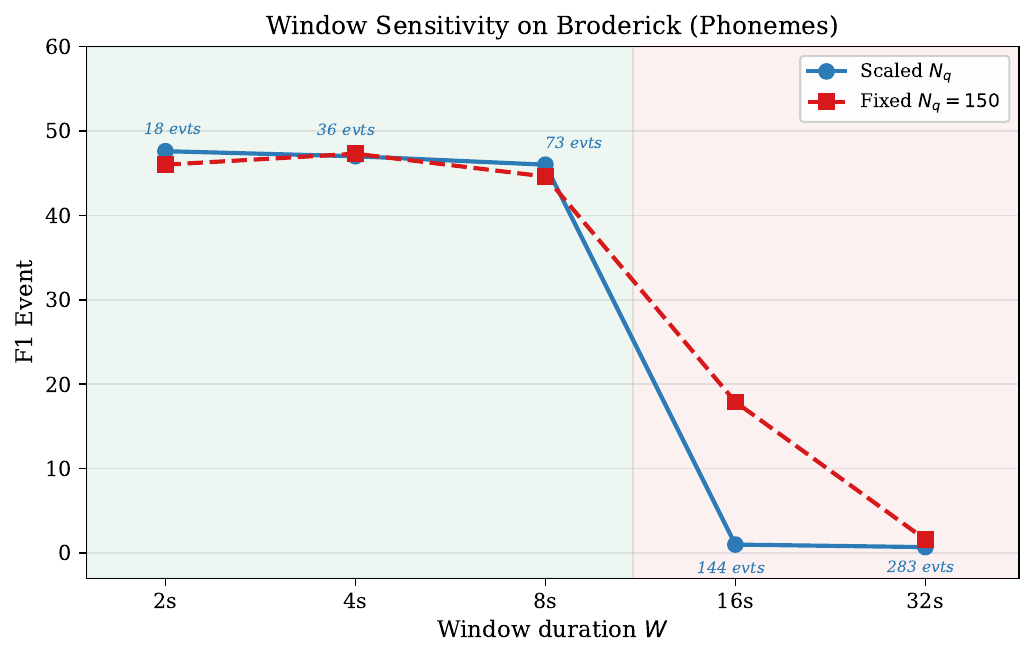}
    \caption{\rev{\textbf{Window size is determined by the interplay of
    coverage and resolution.} Window-size sweep on the phoneme dataset. The solid
    blue curve scales the number of event queries with $W$; the dashed
    red curve keeps $N_{Q_{\text{event}}} = 150$ fixed. Blue annotations
    report the mean number of events per window at each $W$. The green
    band marks the regime where the Perceiver resolves each event with
    at least three latent tokens ($256 / \bar{n} \geq 3$); the red band
    shows where resolution drops below this threshold and
    $F1_{\mathrm{event}}$ collapses \emph{regardless} of $N_{Q_{\text{event}}}$.
    $W = 8$\,s, used in the main experiments, is the largest window
    that remains inside the green regime for this dataset.}}
    \label{fig:window_size_constraints}
\end{figure}

\rev{\Cref{fig:window_size_constraints} validates this design on
the phoneme case, the dataset closest to the resolution bound: at $W = 8$\,s
($3.4$ tokens/event) \model{} is at its optimum, while doubling $W$
pushes $\bar{n}$ past the three-token cutoff and $F1_{\mathrm{event}}$
collapses regardless of how queries are allocated. The fixed-$N_{Q_{\text{event}}}$ baseline falls to $18\%$, and scaling $N_{Q_{\text{event}}}$ in proportion to $W$ drops further to $\sim 1\%$. }

\rev{\Cref{tab:window_resolution} shows that every other dataset sits comfortably inside the safe regime, with Tokens/Event ranging from $4.3$ on BI2013 up to $1024$ on TUSZ. This gives a clear heuristic on how to find the grid of valid $(W, N_{Q_{\text{event}}})$.}

\begin{table}[ht]
\caption{\rev{Window configuration and Perceiver resolution across all datasets. The Perceiver module resamples every window into 256 latent tokens regardless of duration. Tokens/Event is computed as $256 / \bar{n}_{\text{events}}$ and must remain $\geq 3$ for accurate boundary regression. Broderick (3.4) sits at the empirically validated threshold; all other datasets have substantial headroom.}}
\label{tab:window_resolution}
\centering
\small
\renewcommand{\arraystretch}{1.2}
\begin{tabular}{l c c c c}
\toprule
\textbf{Dataset} & \textbf{W (s)} & \textbf{Events / Win} & \textbf{Event Dur.\ (s)} & \textbf{Tokens / Event} \\
\midrule
Broderick    & 8   & 75.3  & 0.08  & 3.4  \\
SparrKULee   & 8   & 19.5  & 0.30  & 13.1 \\
Pinet        & 8   & 19.3  & 0.11  & 13.3 \\
BI2013       & 32  & 60.1  & 1.00  & 4.3  \\
BI2014a      & 32  & 45.8  & 1.00  & 5.6  \\
BI2014b      & 8   & 9.9   & 1.00  & 25.9 \\
BNCI2014     & 8   & 1.3   & 4.00  & 197  \\
BNCI2015     & 32  & 3.4   & 5.00  & 74.4 \\
TUAR         & 16  & 11.0  & 9.04  & 23.2 \\
TUSZ         & 256 & 0.25  & 67.7  & 1024 \\
\bottomrule
\end{tabular}
\end{table}

\rev{\subsection{Qualitative comparison on motor-imagery datasets}}
\label{app:decoded_examples_bnci}

\begin{figure}[h]
    \centering
    \includegraphics[width=\linewidth]{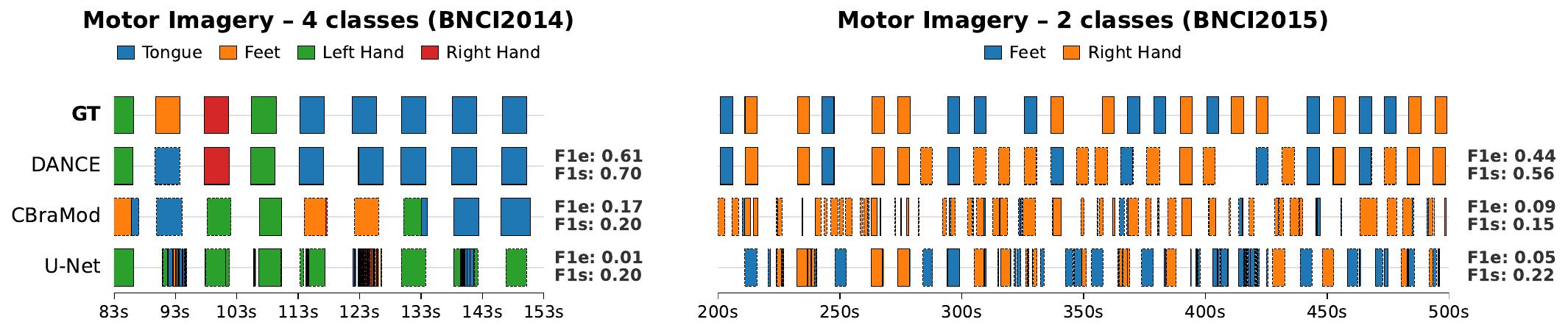}
    \caption{\textbf{\rev{Qualitative timelines on motor-imagery datasets.}}
    \rev{One row per method (top: ground truth; below: DANCE, CBraMod,
    U-Net). Coloured rectangles encode the predicted event class;
    solid borders mark predictions matching a ground-truth event
    (class and IoU $\geq 0.5$), dashed borders mark mismatches.
    Left: BNCI2014 (4 classes), 83--153\,s of one subject.
    Right: BNCI2015 (2 classes), 200--500\,s of one subject.
    Per-row $F1_{\mathrm{event}}$ (F1e) and $F1_{\mathrm{sample}}$ (F1s) on this window are reported on the right.}}
    \label{fig:decoded_examples_bnci}
\end{figure}

\rev{Figure~\ref{fig:decoded_examples_bnci} shows decoded timelines of DANCE, CBraMod and U-Net on a single window of each motor-imagery dataset. DANCE preserves the coherent block structure of the ground truth, while CBraMod (foundation backbone with a dense head) misclassifies events and U-Net (sequence-labeling) produces the fragmented predictions characteristic of dense classifiers: the two failure modes that end-to-end set-prediction is designed to avoid.}

\section{Details on Methods}
\label{sec:appendix_methods}

\subsection{Annotations, Label Standardization and Predictions}
For eight of the ten datasets, annotations are used in their original format. However, the Temple University datasets (TUAR \citep{hamid2020} and TUSZ \citep{shah2018}) require specific handling as they provide channel-specific annotations. Since our framework targets global event detection, we project these annotations onto a single global timeline. Specifically, we merge simultaneous channel-level events, retaining a single event instance per start time per session to eliminate redundancy.

We also standardize the label spaces across tasks. For TUAR (artifacts), we define five distinct artifact classes: \textit{muscle (musc), eye movement, electrode pop (elpp), chew,} and \textit{shiver}, with composite events duplicated and reassigned to their corresponding classes. For TUSZ (seizures), annotations are mapped to four clinical seizure types: \textit{focal, generalized, complex partial (CT),} and \textit{absence}, following previous works \citep{seizure_class}.

For the baseline models and the \model{} w/out Decoder variant, which rely on dense temporal predictions, we derive discrete events by extracting contiguous segments from the timestep prediction masks, resulting in a list of tuples defined by $(start, end, class)$. In contrast, for the \model{} w/out Hybrid ablation and the full \model{} architecture, the decoder head directly regresses these parameters, outputting explicit start coordinates, end coordinates, and class labels for each query within a window.

\subsection{Post-processing Evaluation for Baseline Models}
\label{sec:appendix_gap_filling}
\begin{figure}
    \centering
    \includegraphics[width=0.85\linewidth]{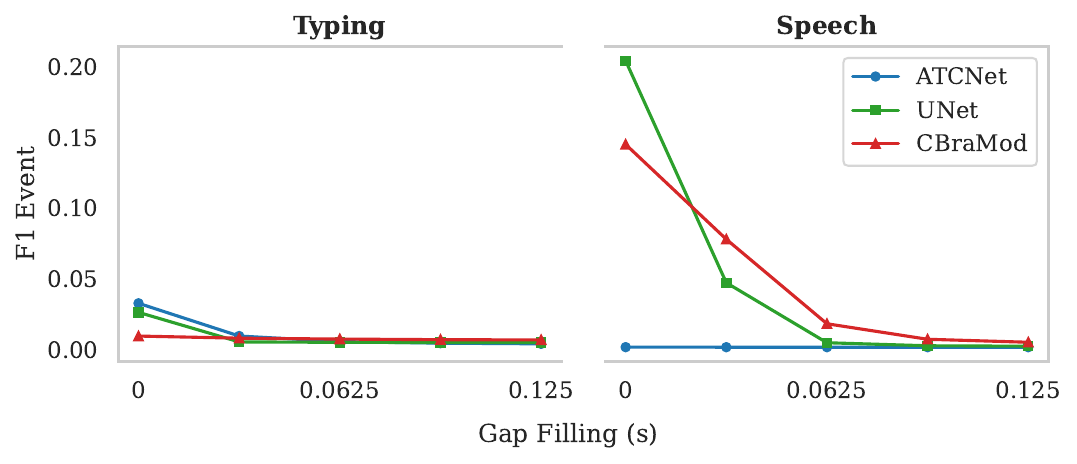}
    \caption{
    \textbf{Temporal smoothing fails to improve event-based performance for baseline architectures.}
    The impact of the gap-filling parameter $\tau$ on the $\mathbf{F1_{\text{event}}}$ score for baseline methods (ATCNet, U-Net, CBraMod).
    For each $\tau \in [0, 0.125]\,$s (x-axis), disjoint predictions of the same class are merged into a single continuous event if separated by a gap of $\leq \tau$.
    Results are shown for the Pinet (left) and Broderick (right) datasets.
    }
    \label{fig:gap_filling}
\end{figure}

Unlike standard object detection pipelines, our proposed method operates without non-maximum suppression (NMS) or duplicate-removal steps; predictions are directly interpreted as events.

To ensure a rigorous comparison, we investigated whether the baseline architectures (ATCNet, U-Net, CBraMod) would benefit from post-processing steps. As these models output dense, timestep-level probabilities, they are susceptible to prediction fragmentation, where continuous events are split into disjoint segments. We evaluated a gap-filling strategy where same-class segments separated by $\leq \tau$ are merged. The parameter range $\tau \in [0, 0.125]\,$s was chosen to encompass durations up to the approximate mean length of typical speech and typing events. This range specifically targets the smoothing of short, spurious discontinuities.

As shown in \Cref{fig:gap_filling}, this heuristic did not yield consistent improvements in $\mathbf{F1_{\text{event}}}$. In many instances, increasing the temporal threshold $\tau$ degraded performance by erroneously merging distinct adjacent events.

While other post-processing steps exist, a central objective of this study is to validate the feasibility of a truly end-to-end pipeline that eliminates the dependency on manual intervention. The process of tuning post-processing stages introduces a selection of arbitrary thresholds or merging rules that adds significant complexity and risk of overfitting to specific datasets. Consequently, to ensure a strictly parameter-free evaluation and a transparent comparison, we report all baseline results based solely on raw model predictions.

\subsection{Model Configuration, Losses and Training}
\label{sec:general_appendix_details}

\subsubsection{Matching versus Evaluation}

During training, the model optimization relies on the set-based loss without thresholding; instead, we utilize the Hungarian algorithm to enforce a global one-to-one assignment between the full set of model queries and the ground truth. Consequently, the majority of queries are matched to the background class to account for the sparsity of ground truth events relative to the query volume.\\
Conversely, during evaluation, we select the class that yields the highest probability. A converged model effectively suppresses irrelevant queries by assigning them high probability in the background class. False positives are defined as non-background predictions that are incorrect, while false negatives correspond to ground-truth events unmatched by any valid foreground prediction.

\subsubsection{Classification and IoU Losses}
The classification loss $\mathcal{L}_{cls}$ is defined as the standard cross-entropy between the predicted probabilities $\hat{p}$ and the ground truth class $c$:
\[
\mathcal{L}_{cls}(c, \hat{p}) = - \sum_{k} \mathbf{1}_{[c = k]} \log \hat{p}_k
\]

The IoU loss $\mathcal{L}_{iou}$ for 1D temporal intervals with predicted $(p_s, p_e)$ and target $(t_s, t_e)$ is:
\[
\text{IoU} = \frac{\max(0, \min(p_e, t_e) - \max(p_s, t_s))}{\max(10^{-6}, \max(p_e, t_e) - \min(p_s, t_s))}, \quad
\mathcal{L}_{iou} = 1 - \text{IoU}.
\]

Standard set-prediction architectures typically rely on Generalized IoU \citep{Rezatofighi2019GeneralizedIOU} to avoid vanishing gradients when predictions do not overlap with the ground truth. In our case, however, dense auxiliary supervision provides sufficient signal, making standard Intersection over Union adequate.

These losses are combined in the matching cost as:
\[
\mathcal{L}_{match} = \lambda_{cls} \mathcal{L}_{cls} + \lambda_{iou} \mathcal{L}_{iou}
\]

\subsubsection{Consistency Loss Formulation}
\label{sec:appendix_consistency}
The consistency loss $\mathcal{L}_{cons}$ bridges the gap between the continuous dense stream predictions and event proposals. To compute it, we unfold the sparse event predictions into a dense map $P_{event} \in \mathbb{R}^{N_{time} \times K}$. This projection is used only as a probabilistic target and is differentiable solely with respect to the event class probabilities, not the predicted temporal boundaries. As such, the loss provides supervision to the semantic content of event predictions while leaving temporal localization unaffected.

First, for every timestep $t$ in the downsampled sequence, we compute an unnormalized score $\tilde{P}_{event}(t, k)$ by aggregating the probability distributions of all predicted events covering that instant:
\begin{equation}
    \tilde{P}_{event}(t, k) = \sum_{i=1}^{N_{Q_{\text{event}}}} \mathbf{1}_{[t \in \hat{b}_i]} \cdot \hat{p}_i(k)
\end{equation}
where $\hat{p}_i(k)$ denotes the softmax probability of class $k$ for the $i$-th query, and $\mathbf{1}[t \in \hat{b}_i]$ is the indicator function, equal to 1 if $t$ falls within the predicted interval $\hat{b}_i$.

We then normalize these accumulated scores to obtain a valid probability distribution $P_{event}(t, k)$, introducing a small constant $\epsilon$ for numerical stability in regions devoid of event predictions:
\begin{equation}
    P_{event}(t, k) = \frac{\tilde{P}_{event}(t, k)}{\sum_{j=0}^{K} \tilde{P}_{event}(t, j) + \epsilon}
\end{equation}

Finally, $\mathcal{L}_{cons}$ introduced in \Cref{sec:consistency_loss} applies a Kullback--Leibler (KL) divergence loss between the dense temporal probability distribution $P_{dense}$ and the unrolled event-derived probability map $P_{event}$.

We investigated replacing $P_{dense}$ with the dense ground-truth labels and further evaluated a fully differentiable formulation of the consistency loss designed to propagate gradients also to the localization components. However, neither configuration outperformed the alignment of the $P_{dense}$ and $P_{event}$ probability distributions. This can suggest that the consistency term may act as a form of auto-distillation \citep{hinton2015distilling}, where the divergence between predictions provides a richer regularization than hard labels alone.

\begin{table}[ht]
\caption{Additional information on the experimental datasets.}
\label{tab:datasets_details}
\centering
\small
\renewcommand{\arraystretch}{1.2}
\setlength{\tabcolsep}{6pt}

\begin{tabular}{l l p{4.5cm} c c}
\toprule
\textbf{Task Domain} & \textbf{Dataset} & \textbf{Classes ($K$)} & \textbf{Window (s)}  & \textbf{Queries} \\
\midrule
\multirow{2}{*}{\textbf{Motor Imagery}}
 & BNCI2014   & L. Hand, R. Hand, Tongue, Feet  & 8  & 5  \\
 & BNCI2015 & R. Hand, Feet     & 32 & 40   \\
\midrule
\multirow{2}{*}{\textbf{Speech}}
 & Broderick   & Phonemes (Voice Detection)   & 8  & 150  \\
 & SparrKULee  & Words (Voice Detection)      & 8  & 50  \\
\midrule
\multirow{3}{*}{\textbf{P300}}
 & BI2013  & Target vs.\ Non Target    & 32 & 150 \\
 & BI2014a & Target vs.\ Non Target  & 32 & 150  \\
 & BI2014b & Target vs.\ Non Target & 8  & 150 \\
\midrule
\textbf{Seizure} & TUSZ (2.0.3) & CT, Focal, Generalized, Absence  & 256 & 10  \\
\midrule
\textbf{Artifacts} & TUAR & Chew, Elpp, Eye Mov, Musc, Shiver & 16 & 25  \\
\midrule
\textbf{Typing} & Pinet & Key Press & 8 & 100  \\
\bottomrule
\end{tabular}
\end{table}

\subsubsection{General Implementation Details}
\label{sec:reproducibility}

\Cref{tab:datasets_details} summarizes the dataset-specific hyperparameters. Across datasets, only three parameters vary: the window size, the number of queries and the batch size. 
\rev{We have defined in \Cref{sec:window_size_constraints} how to select the range for the window size. The number of queries must exceed the maximum number of events per window to guarantee full detection. Because excess queries simply predict "no-object" and linearly increase the bipartite matching cost during training, we set this parameter to a moderate multiple of the expected event count to balance coverage and efficiency. The batch size is chosen to maximize GPU utilization depending on the selected window size.}

%These values were determined via a systematic grid search optimization. The window size $T$ was evaluated over a logarithmic scale (powers of two), bounded strictly below by the maximum observed event duration and above by hardware memory constraints. We also ensure a viable batch size $B \in [2, 128]$. The number of queries $N_{Q_{\text{event}}}$ was optimized by evaluating candidate values that strictly exceed the maximum event density for a given window $T$, guaranteeing sufficient capacity to capture all potential events.

To ensure convergence and prevent overfitting, the convolutional backbone employs Batch Normalization coupled with GeLU activations for non-linearity. Regularization is managed via a dropout schedule, where a probability of $p = 0.1$ is applied to the raw input, while a rate of $p = 0.2$ is used within the convolutional blocks.

The complete architecture composed of the convolutional backbone, Perceiver, and the decoder is trained end-to-end. The model contains approximately 31\,M parameters (27\,M allocated to the encoder/perceiver and 4\,M to the decoder). We use the AdamW optimizer \citep{loshchilov2017decoupled}. Unless otherwise specified, we train for 100 epochs with early stopping (patience = 20). We employ a OneCycle learning rate scheduler \citep{smith2017super}, with a warm-up phase covering the first 10\% of training. The maximum learning rate is set to $5 \times 10^{-5}$ for all datasets with the exception of the TUAR dataset, which required a higher rate of $10^{-4}$ to ensure convergence with our method.

$\mathbf{F1_{\text{sample}}}$ was computed using the \texttt{MultilabelF1Score} implementation from \texttt{torchmetrics}, while $\mathbf{F1_{\text{seizure}}}$ was computed using \texttt{BinaryF1Score}.

\subsubsection{Seizure-Specific Optimization}
Seizure detection presents unique challenges due to class imbalance; there is approximately 20 times more background than seizure activity in this dataset. To address this and inspired by \citet{wu2025seizuretransformer}, we employ a weighted sampling strategy based on window occupancy. Let $o_i \in [0, 1]$ be the percentage of the window $i$ containing seizure activity. The sampling weight $w_i$ is calculated as:
\begin{equation}
    w_i = \alpha + (1 - \alpha) \cdot \min(1, \beta \cdot o_i)
\end{equation}
where $\alpha = 0.01$ ensures a minimum probability for background windows, and $\beta = 5$ boosts the sampling of sparse seizure events.
We also introduce a 50\% overlap between consecutive windows at training time to ensure consistent supervision for events spanning window borders. The model is trained for 25 epochs without early stopping. Importantly, with a duration $= 256\,$s, the final window of each session is padded to the full window size, ensuring that inference can be performed over the entire test set. In \Cref{fig:seizure}, we report the averaged performance across 3 model seeds on the fixed test set to account for variance between runs.

\subsubsection{Compute Resources}
\label{sec:compute_resources}

\rev{\model{} has an average runtime of 8.03\,h (training and evaluation) and an inference latency of 34.5\,ms per single-window input. While higher than CBraMod (5.56\,h, 28\,ms), \model{} provides more robust performance and maintains the potential for real-time decoding with a limited increase in inference time.}

\begin{table}[!h]
\centering
\small
\setlength{\tabcolsep}{4pt}
\begin{tabular}{lcccccc}
\toprule
 & \textbf{ATCNet} & \textbf{U-Net} & \textbf{CBraMod} & \textbf{\model{} w/out Decoder} & \textbf{\model{} w/out Hybrid} & \textbf{\model{}} \\
\midrule
\textbf{Runtime (h)} & 1.18 & 1.14 & 5.56 & 3.51 & 5.85 & 8.03 \\
\textbf{Latency (ms)} & 2.35 & 4.65 & 28.02 & 34.81 & 34.03 & 34.53 \\
\bottomrule
\end{tabular}
\caption{Average runtime and inference latency across all datasets for each method.}
\label{tab:runtime_summary}
\end{table}

All models were trained individually for each dataset on a single NVIDIA V100 GPU, except for the seizure detection task. Due to its larger scale, this dataset was trained using distributed data parallelism across 8 GPUs. \Cref{tab:runtime_summary} reports the computational performance averaged across the 10 datasets: \emph{runtime} corresponds to the total pipeline duration (training and evaluation), while \emph{latency} represents the time required to process a single-window input during inference.

It should be noted that the metrics reported above reflect only the final, converged evaluation runs. The complete lifecycle of this research project required more computational resources. Developing the \model{} architecture, conducting hyperparameter sweeps to find a robust configuration across ten highly heterogeneous datasets, and evaluating preliminary designs that were ultimately discarded accounted for the majority of the total computational budget. 

\subsection{Evaluation}
\label{sec:appendix_evaluation}

\subsubsection{Splitting and Aggregation}
We adopt a subject-independent evaluation protocol to ensure the generalization capability of the models.
\begin{itemize}
    \item \textbf{General Protocol:} For most datasets, we employ a 5-fold cross-validation scheme grouped by subject. This approach ensures a complete partition of the dataset, effectively implementing a leave-N-subjects-out evaluation where each subject is tested exactly once. Metrics are first computed individually for each subject. Consequently, the results reported are the means across all subjects (\Cref{fig:performance}).
    \item \textbf{Seizure Detection:} We adhere to the official train/test split defined by the TUSZ dataset. Performance is evaluated on the entire held-out test set.
\end{itemize}

Regarding the highly structured protocols of the BNCI datasets, we ensured that the effective batch size defined as the product of batch size (BS) and window length (W) remains negligible relative to the total session duration. It is restricted to 64\,s for BNCI2015 ($BS=2, W=32$\,s) and 32\,s for BNCI2014 ($BS=4, W=8$\,s), very far from the mean session durations of 35.76\,min and 6.45\,min, respectively. This gap mitigates the risk of the model overfitting to macro-scale protocol regularities or memorizing absolute temporal positions within a session.

\subsubsection{Statistical Analysis}
\label{sec:statistics}
We assessed statistical significance using Linear Mixed-Effects Models (LMMs), with subjects and datasets modeled as random effects to account for repeated measurements and domain-specific variability. This approach ensures that statistical significance reflects consistent effects across subjects and domains, rather than being driven by sample size imbalances or over-representation of specific datasets. Linear mixed-effects models were implemented using the \texttt{statsmodels} Python library \citep{seabold2010statsmodels}.

\subsubsection{Baselines and Ablations}
\label{sec:appendix_baselines}
We benchmark our approach against a diverse set of methods ranging from random baselines to state-of-the-art deep learning architectures.

\paragraph{Note on Architecture Selection.} We exclude standard classification models tailored for single-window inputs, such as EEGNet \citep{lawhern2018eegnet}. Adapting such architectures for temporal detection requires a sliding-window approach; however, a stride-1 evaluation leads to computationally intractable inference times ($>24$ hours) on large-scale datasets. Consequently, we restrict our scope to architectures that can be modified or are designed for dense prediction.

\paragraph{Random Baseline.} To establish a non-trivial lower bound, we generate random predictions by sampling directly from the list of all ground truth events in the dataset. For each session, the number of events injected is determined by the mean event rate in the dataset, while their start times are assigned uniformly at random.

\paragraph{ATCNet \citep{altaheri2022physics}.}
We use ATCNet, an attention-based convolutional network, adapted here for dense prediction tasks. Our implementation utilizes the \texttt{braindecode}\footnote{\url{https://braindecode.org/stable/index.html}} library \citep{braindecode}. To enable dense sequence labeling, we modified the architecture to preserve the input's temporal resolution. Specifically, we disabled temporal downsampling by setting the pooling parameters \texttt{conv\_block\_pool\_size} to 1 and treated the input as a single continuous sequence by setting \texttt{n\_windows} = 1. Furthermore, we retained the full output sequence of the Temporal Convolutional Network (TCN) rather than extracting only the final timestep. All other hyperparameters were maintained at their default values.

\paragraph{U-Sleep \citep{perslev2021u}.}
Given the dominance of U-Net architectures in the segmentation literature, we include the U-Sleep model \citep{perslev2021u} as a strong baseline, utilizing the implementation provided by \texttt{braindecode}. To accommodate shorter input windows and prevent excessive feature collapse, we reduced the network depth from 12 to 8. Additionally, we modified the classification head by removing the adaptive pooling layer, ensuring the model produces a dense representation for every timestep.

\paragraph{LaBraM \citep{labram}.}
\rev{LaBraM is a transformer-based EEG foundation model pretrained on
$\sim$2{,}500 hours of clinical and consumer-grade EEG. Pretraining
follows a masked-token reconstruction objective: 1-second EEG patches
are encoded by a vector-quantised tokenizer and the transformer is
trained to recover the discrete tokens of randomly masked patches from
their context, with shared per-channel embeddings allowing the same
backbone to ingest heterogeneous electrode montages. We use the released
pretrained weights and default configuration, and replace the final
linear head with a multi-layer perceptron that produces a dense
per-time-step class distribution.}

\paragraph{REVE \citep{elouahidi2025reve}.}
\rev{REVE is a more recent EEG transformer pretrained on a
$\sim$60{,}000-hour multi-source corpus mixing clinical, BCI and
consumer-grade recordings. It's one order of magnitude larger than the
corpora used by LaBraM or CBraMod. The architecture stacks $22$
transformer blocks at hidden dimension $512$, making it the largest of
the three foundation backbones we evaluate. We follow the same protocol
as for LaBraM and CBraMod.}

%All baselines were optimized via grid search over learning rate and batch size.

\rev{Every baseline underwent a systematic search for optimal learning rates and batch sizes. For the dense baselines, we tested decision boundaries and gap-filling thresholds to maximize their temporal coherence. All models were required to converge across the 10-dataset benchmark using a unified configuration. This strict constraint prevents hyperparameter over-optimization.}

\subsubsection{Ablations}
\label{sec:appendix_ablations}
We conduct an ablation study to isolate the contributions of the proposed architecture and learning objectives:

\begin{enumerate}
    \item \textbf{\model{} w/out Decoder:} This variant removes the DETR-style decoder and trains the backbone solely using the dense cross-entropy loss. This tests the efficacy of the encoder architecture in isolation.
    \item \textbf{\model{} w/out Hybrid:} This variant includes the full architecture but removes the dense and consistency auxiliary losses from the supervision. This highlights the value of our hybrid training strategy in aligning the discrete (event) and continuous (signal) latent representations.
\end{enumerate}

\section{Broader Impacts}
\label{sec:broader_impacts}
\model{} is a unified framework that detects and classifies heterogeneous EEG events from continuous recordings, bypassing the unrealistic reliance on known event timings. By combining set-based prediction with dense supervision, it provides a scalable, deployment-faithful path for asynchronous EEG decoding. Regarding broader impacts, this work advances positive clinical applications—such as automated seizure monitoring to enhance patient care—and asynchronous brain-computer interfaces for individuals with severe motor or communication impairments. Conversely, improved continuous decoding may introduce societal risks. In clinical settings, it could encourage automation bias; \model{} should support, not replace, expert medical judgment. Additional concerns include mental privacy and the potential surveillance of cognitive states, which warrant careful oversight. Because this work relies exclusively on public datasets, it does not introduce new data privacy or immediate misuse risks.
\looseness=-1

% =========================================================================
% TODO (rebuttal integration, week of May 2026):
%
% The following sections collect the reviewer-driven analyses produced
% during the ICML rebuttal. They live in scratch/jarod/dance/paper/ as
% LaTeX snippets and figures and are imported here so they can be moved
% into the main body or kept in the appendix as space allows.
%
% \section{Sensitivity to the IoU Threshold}
% \input{reviewer_table}                                  % decoupled F1_detect / F1_event / Δctr
% \includegraphics{figures/fig_f1_event_sensitivity}
%
% \section{Sensitivity to Loss Weights}
% \includegraphics{figures/fig_loss_sensitivity}
%
% \section{Window and Perceiver Resolution}
% \input{table_window_resolution}
% \includegraphics{figures/fig_broderick_window_sensitivity}
% \includegraphics{figures/fig_nqueries_ratio_validation}
%
% \section{Foundation Models with Set Prediction}
% \input{foundation_table}
%
% \section{Onset-informed Comparison (EEGSym)}
% \includegraphics{figures/fig_eegsym_vs_dance}
%
% \section{Refined Decoded-Example Timelines (BNCI)}
% \includegraphics{figures/figure3_decoded_bnci_v3}
% =========================================================================

\end{document}